# M-GWAP: An Online and Multimodal Game With A Purpose in WordPress for Mental States Annotation

Fabio Paolizzo. Department of Cognitive Sciences, University of California, Irvine.

**Abstract:** *M-GWAP is a multimodal game with a purpose of that leverages on the wisdom of crowds phenomenon for the annotation of multimedia data in terms of mental states. This game with a purpose is developed in WordPress to allow users implementing the game without programming skills. The game adopts motivational strategies for the player to remain engaged, such as a score system, text motivators while playing, a ranking system to foster competition and mechanics for identify building. The current version of the game was deployed after alpha and beta testing helped refining the game accordingly.*

## I. INTRODUCTION

Before the Internet era, aggregating information for computational modelling was particularly challenging because a considerable amount of time and effort was required to categorize the information efficiently (Zhu, 2005). This was particularly true for emotions and music (Trohidis et al., 2008; Turnbull et al., 2008), as music implies a great level of subjectivity and cultural implications encompassing perception and cognition. Such tasks became more affordable with Internet Research (Miller, 2008) making it possible to investigate the approach to a multitude of contexts, from music recommendation systems to computational music creativity. Still, data interpretation required the work of experts for making sense of statistically aggregated information. An important turning point in this was achieved by proving that crowd-related problems involving cognition, coordination and cooperation can be solved even more efficiently by leveraging on a phenomenon known as *wisdom of crowds*. Accordingly, crowds exhibit wisdom when four conditions are met: diversity of opinions, independence in providing the personal opinions, aggregation of personal judgments for taking collective decisions, and diversification of the individual areas of expertise (Surowiecki, 2005). Such approach leverages on crowdsourcing for solving large-scale computational problems. However, making people to work cognitively, in a cooperative and coordinated manner, is a task that can turn to be either biased by population pooling and social influence as people always have motivations guiding their own decisions, or particularly expensive for employing people in such a large number as solving computational tasks requires. Notably, the phenomenon can be used effectively in so-called *games with a purpose* (GWAPs), which allow a large base of individuals, who exhibit diversity of opinions and expertise each independently from the other, to classify large amounts of data while engaging them in a game-like experience. That is, a GWAP can leverage on crowdsourcing to solve a research problem by turning the task at hand into a game that can be fun to play. This allows focusing wisdom of the crowds on very complex tasks while containing cost considerably (e.g., Sullivan et al., 2018).

In the present study, we focus our research on developing a method for aggregating information that can help in closing the gap which separates computational from human musical creativity. Music cognition is an embodied process where listeners semantic relations with the sonic world can change through functional adaptation at the level of sensing, acting and coordinating between action and perception, in biological, psychological, and cultural terms that involve motor, kinesthetic, haptic and visual, besides the purely auditory components (Reybrouck & Eerola, 2017). Advancing research in the modelling of human musical creativity can be possible by enabling systems to classify, recognize and predict emotional states in experiences of "musicking", which encompass domains such as the sonic, visual, motion and linguistic, at the very least. An approach to the modelling of human music creativity is that of enabling in computational creativity systems the capacity to take decisions by combining information from different modes of sensing. This suggests that such systems could draw on analogies from other forms of knowledge like humans regularly do in music making.

Although the use of GWAP methods leveraging on wisdom-of-the-crowds approaches are more efficient than differently based computational means of learning (Von Ahn & Dibbish, 2008), the scenario has changed in recent years. Unsupervised learning is compelling in approaches to classification (Ngiam et al., 2011; Radford et al., 2015; LeCun et al., 2015; Conneau et al., 2017; Schmidhuber, 2015). Computational creativity as a field of study has delivered numerous systems that fall in the category either of co-creation or of autonomous creation (Ventura, 2017). Therefore, adopting wisdom-of-the-crowds approaches may appear debatable in those fields. Still, human beings can perform tasks that a computer is still completely unable to perform. In creative practice, the dealing with the complexity of human emotions is a common task. Creative practice constitutes a closer approximation to real-world scenarios in comparison to the sole use of labels for describing of musically induced/expressed emotions. If we want to understand how we feel and attribute meaning when interacting with music technology, we can better do so through methodologies involving creative practice at some level. Implementing such understanding can provide next-generation systems of computational music creativity to operate in terms that can represent human cognition as multimodal and embodied. We expect the modelling of how humans "make analogies" in accordance to how they feel to be validated successfully by combining results from wisdom-of-the-crowds approaches with taxonomies derived from case studies that involve involving motor, kinesthetic, visual and language besides auditory components. New interactive and/or intelligent audio (Paolizzo & Johnson, 2017) (Surges & Dubnov, 2013; Surges et al., 2016) and music systems (Assayag et al., 2006; Lévy et al., 2012, p. 1) (Rivaud & Pachet, 2018) (Sturm et al., 2019) could leverage on such new methods to advance beyond the current state of the art. On a broader scale, these methods can enable advancements in the fields of information retrieval, artificial intelligence and computational creativity.

We propose a GWAP approach for the computational classification of induced emotions, which can be expressed through musical media data (Paolizzo et al., 2017). The present approach is functional to the development of an online database of scores, lyrics and musical excerpts, vector-based 3D animations, and dance video recordings, indexed by mental states (Paolizzo, 2019). M-GWAP is a multimodal game with a purpose for Internet users that can be used to generate ground truth for the classification of multimedia in terms of mental states. In section A, we consider the problem of advancing research in computational music creativity through the computational modelling of emotions. Section B presents GWAP research which can be of interest for investigating that problem. Section X of the present paper describes the design of M-GWAP as a method to solve our goal. Section Y presents practical experiments resulting from participants' engagement in M-GWAP tests in a lab environment (Paolizzo at al., 2017). Section Z considers our findings for increasing interaction capacities and/or computational intelligence by the means of information retrieval techniques.

## II. CLASSIFICATION OF EMOTIONS

Mental states are the result of emotional and cognitive processes. The capacity to make analogies that involve multimodality and embodiment, necessarily involves emotions at same level. We can recall three types of emotional analogies, such as 'analogies and metaphors about emotions […], analogies that involve the transfer of emotions […], analogies that generate emotions' (Thagard and Shelley, 2001). Because we can set up a system to interact with a user, and observe their interaction, IMSs offer the opportunity for investigating how we as humans translate between the sonic, visual, motion and language domains. We can model human emotions for computational music creativity in ways that involve cross-modal computational means of representing the process of "making analogies". Advancing IMSs is therefore possible by implementing in such systems a capacity for "embeddedness", that is the capacity of a system to change its own semantic relations to/with the sonic world in biological, psychological and cultural terms, in a bidirectional circle of action, perception and reaction between user and system. This has been investigated in various terms, as the action-reaction cycle (Leman, 2008; Keebler et al. 2014), the action-perception feedback loop (Vaggione, 2001), or the analysis of semiosis in the human/music interface (Stefani, 1998), as well as for IMSs specifically (Paolizzo & Johnson, 2017; Paolizzo, 2017). IMSs can benefit from wisdom-of-the-crowds approaches for drawing cross-modal references between emotions experienced by individuals and different modes of their perception.

Computational models of meaning based on wisdom of the crowds provide the highest accuracy when the classification of meaning is grounded in perception (Bruni et al., 2014). A first and simple approach for classifying music in terms of emotions is that of having a listener indicating what is the emotion expressed by a set of music stimuli. Because annotators self-report the music induction from listening sessions in accordance to their subjective experience, we can assimilate the task of music emotion labelling to a process of meaning attribution that is based on the perception of emotions. Emotion induction through music is therefore the process of emotionally affecting a subject through musical stimuli that involve biological, cognitive and cultural implications. However, various difficulties in emotion representation can be encountered. These may include the annotator's subjectivity and the language constraints. Different subjects can simply have a different opinion about the same emotional content. The need for a universal model able to represents real-world scenarios is paramount. A solution to this is that of considering multiple annotations to describe the same emotion for each piece of music in a dataset; different annotators can annotate a subset of the dataset through predefined tags. This is known as multiclass annotation (Aljanaki et al., 2014) (Skowronek et al., 2007). A more nuanced approach capable of better modelling emotion induction is that of using continuous values representing different dimensions of the perceived emotions, which are attributed to a specific music piece (Laurier et al., 2009) (Yang et al., 2009) , typically adopting the valence-arousal Cartesian expressive space (Kim et al., 2010). These dimensions are independent, with valence indicating how a person feels depending on positive or negative evaluations of people, things or events, and arousal indicating the degree of a person's activation and his/her inclination to perform actions. It is possible to extend the valence-arousal space by a third dimension as discussed in MIR-based psychology (Schimmack & Reisenzein, 2002).

Emotions are better understood and captured when observed in their multimodal complexity. For example, a same music piece can channel single or multiple emotions simultaneously. We approach the scenario of emotions expression/induction through music as a multilabel and multiclass problem, where multiple emotion labels can be adopted for the same music by each annotator (multilabel), and each emotion can be identified or not in the music (multiclass) simultaneously. We consider different distributions of annotations and emotion labels in a corpus by considering emotion labels as valid labels for a music piece when the mean positive response of annotations per label surpasses a specific consensus threshold. A first consideration can be drawn regarding the use of consensus thresholds as a means of validation for multiple emotion labels to identify the number of available labels used for describing effectively induced/expressed emotions through music. The high value of mean performance achieved by classifiers (Paolizzo et al., 2019) confirms the validity of consensus threshold as a method for the analysis of the distribution of annotations per emotion label. In designing a GWAP for the classification of musically expressed/induced emotions, we should therefore aim at gathering enough annotations that can represent a consensus for emotion labels proportional to the datapoints distribution that is equal or higher to threshold values adopted. In our parallel study, we have considered that at least 25% – 30% of the total annotators will need agreeing regarding the identification of a specific emotion in an identical music excerpt, for the emotion label to be valid and the classifiers achieve strong performance. In M-GWAP, scores are based on popularity rankings for emotion labels, as discussed in section X. Popularity is therefore assimilated to a form of consensus. An emotion term needs to pass a threshold of popularity before it is added as the real annotation of a snippet. In the next section, we will review most common types of games with a purpose that can be used for the computational modelling of emotions.

# III. TEMPLATES OF GAMES WITH PURPOSE

GWAPs are a form of crowdsourcing gamification adopted for solving large-scale computational problems. Therefore, user motivation is a key factor in solving the research problem successfully. The users must be able to play the game in an engaging and seamless manner.

We can identify three types of game-structure templates that are suitable to classification of induced emotions, such as output-agreement, inversion-problem and input-agreement games. Table 1 presents these GWAP-structure templates. Each one of them allows randomly choosing two human players among all the players. Each type of game provides players with an input that can be different or equal for both players, or provides an input for one player who is then required to elaborate that input for another player. Once the game provides the input, players must choose or describe among the inputs they have received, in order to produce an output that defines if they have won the game and scored points, or not.

Table 1. GWAP structure templates.

|  | **Output-agreement games** | **Inversion-problem games** | **Input-agreement games** |
|---|---|---|---|
| **Random player** | Yes | Yes | Yes |
| **Matched players** | 2 | 2 | 2 |
| **Input** | Provided by GWAP to players | Provided by GWAP to players (describer). | Provided by GWAP to players for categorization agreement (same/different) |
| **Output** | Players guess among provided inputs | Describer produces output for guessers | Players describe input to each other |
| **Winning condition** | Both players agree on same output | The guesser produces the same input given to the describer. | Both players guess if inputs are same or different |
| **Rewards** | Score | Score | Score |
| **Additional elements** | N/A | Players' action transparency; Asymmetric alternation | N/A |

Increasing the number of users and average time played by each of them is critical in designing a game structure. Desired features include comprehensible and challenging goals (Von Ahn & Dabbish, 2008), appealing design and simplicity of gameplay, appropriate and motivating score system and in-built methods for establishing and extending a player base (Dulacka et al., 2012). With the aim of enhancing game gratification for the user, we incorporate a factor of challenge in the game features. M-GWAP challenges players' factors of engagement such as score keeping, high-score lists, badges and randomness of provided media input. The design structure of GWAP consents users to consult leaderboards for visualizing scores with their own rank information for all games played. For each emotion-label provided, players earn a score according to the answers they provide. Randomness increases challenge by scrambling the difficulty level of media clips received by a player. By doing so, players are in constant challenge, because they are not aware of how difficult the next step of the game will be.

Research has shown that imposing a time constrains in an annotation task can induce a stress condition ultimately affecting annotations themselves. Collecting data through M-GWAP depends on humans that voluntarily play the game. With so many competitors that aim to gain players time and attention, is imperative to make an enjoyable and fun game. In order to avoid players to provide answers that are generic, repetitive or inaccurate in response to such a stress condition, pre-defined emotion labels are not provided. For the same reason, we decided not to impose a time limit on the player.

## IV. M-GWAP: A MULTIMODAL GAME WITH A PURPOSE

The widespread diffusion of data science has progressively extended the user base and included those who have limited knowledge of the technology required to take advantage of wisdom-of-the-crowds approaches for aggregating information online, meaningfully. WordPress is a PHP-based, online and open source tool for website creation, which has become a standard for blogging and content management. The success of the tool is to be found in its ease of use; WordPress does not require programming skills to be used also for complex tasks as the community flourishes with developers who release new modules continuously. In the present article, we include PHP and MySQL implementations of M-GWAP alongside the game features.

In M-GWAP, users must register with their own email and activate the account through a link sent to their inbox, in order to play the game. This ensures that they also have read the study information sheet and provided age and language information, which can be useful in profiling the user base and evaluate the annotations gathered. At the beginning of each game, the player is presented with a 1-to-5 scale for a self-assessment of his/her own mood. In emotion research this information is particularly useful as the mood reported by a subject for a specific snippet could be influenced to the way he/she feels.

In M-GWAP, players are provided with a snippet of content. This can either be text, music or a video. Players respond with text to an emotion label which they think was conveyed by the snippet. More popular is the response to the same snippet in comparison to the other players' responses, more points are assigned. By doing so, users are encouraged to provide a description for describing the emotion which they think could be accepted by most of the other users. Short descriptions, such as single words, are obviously the most popular type of descriptions. This ultimately increase the precision of the annotation task. As soon as a description reaches a threshold of popularity, the string is added to the database as an appropriate annotation for that snippet. This allows increasing the amount of annotations for the snippets in the Musical-Moods database that will be annotated using M-GWAP. In order to facilitate the user experience, design and layout are very simple and goal of the game is very clearly laid out. Instructions are provided to the players requesting them to label the given snippet by the mood it conveys. An in-game tutorial can also be played by all new users and a standard tutorial can be revisited as often as a user wants.

During the game, players receive a randomized snippet, which helps maintaining a certain level of challenge for the user, as mentioned. Players can view the snippet and type in a response. For each snippet labeled by the players, they receive scores. When the database has several entries for a specific mood, players receive a score multiplied by the percentage of players who also provided the same label for that snippet. The score is finally scaled to the number of total responses provided for the snippet.

Motivation is key in gaming. For a starter, players receive 100 points for each new mood-label they provide. The game also shows cheering messages and notes as the player's score increase, which helps maintaining the player's interest high. As M-GWAP is based on the popularity of a mood-label, labels are of high quality when over 90% of the players have provided the same mood-label. When a player provides such a high-quality label, added multipliers increase the score further. At the same time, the game shows a message with the updated score every time a player gains popularity and added multipliers. This way, players receive feedback about their score as well as encouragement to keep trying in thinking like other players. This method has been proven successful in past GWAPs (Dulacka et al., 2012; Aljanaki et al., 2016; Richter et al., 2018). As players can exit the game at any time or continue to play and respond to more snippets, at the end of each game, various motivational messages are shown to players, encouraging them to continue playing. Challenge is a useful method to keep players engaged in the game, as mentioned. For helping to maintain such status on, comparisons among players' ranks are provided. Every player receives a total score aside their current rank in comparison to the others'. The rank is computed by combining the total score of all games played. This way, users are encouraged to play more games, earn more points and increase their rank in order to jump ahead of other players. Table 2 provides a list of all motivators implemented in M-GWAP and which game dynamics they reinforce.

Table 2. List of motivators in M-GWAP.

| Motivator | Description | Game Dynamics |
|---|---|---|
| Score for snippet labeled | Base incentive | Score System |
| Novel label score bonus | Score incentive for new labels | Score System |
| Consensus multiplier | Score multiplied percentage of players with same label | Score System |
| High quality bonus multiplier | Leading players towards most popular mood labels | Score System |
| Cheering messages | Reinforcing player's interest when scoring | Text Motivators |
| Motivational messages | Encouraging players to play again | Text Motivators |
| Snippets random provision | Challenge increased by scrambling difficulty levels | Ranking System |
| End-of-round ranking for players | Displaying total score for players with similar ranking | Ranking System |
| Leaderboard | Fostering competition | Ranking System |
| Badge system | Earning badges while playing | Identify Building |
| User stats pages | Page showing user's total score, highest game score, highest word score, total number of games played and badges won/to win | Identify Building |
| User profile page | Players can use gravatar pictures or upload their own, change personal information (age, spoken languages), profile privacy settings and review the study information sheet | Identify Building |

In order to make the game experience as engaging as possible, while also useful for solving the research task at hand, players receive badges for accomplishments (i.e., reaching a high score, responding to a certain number of snippets during the game, etc.), while playing. Players are encouraged to earn more badges as they continue playing the game. Badges won during the game are also shown at the end of that game along with a list of all badges that can still be won. Each time a badge is earned players can visualize how close they are to earning a new badge, motivating the player in playing the game again. Table 3 provides a list of all badges available in the game and the corresponding threshold for a user to unlock each badge.

Table 3. List of all badges in M-GWAP.

| Badge | Threshold |
|---|---|
| Newbie | Play one game |
| Adventurer | Play 10 rounds |
| 100 Meter Sprint | Respond to 10 snippets in a game |
| Explorer | Play 50 games |
| Marathon Runner | Respond to 50 snippets in a round |
| Precious Gem | Respond with 50 scoring words |
| Special Snowflake | Respond with 50 unique words |
| Crème de la Crème | Make it into the leaderboard |
| Around the World | Respond to 1000 snippets |
| Crème de la Crème | Make it to the leaderboard |
| The Whole Enchilada | Earn every badge |

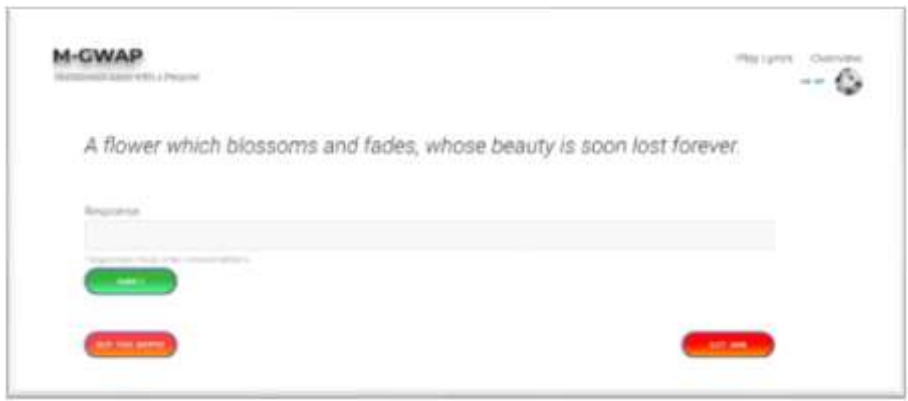

Figure 1: A user is presented with a text snippet.

.

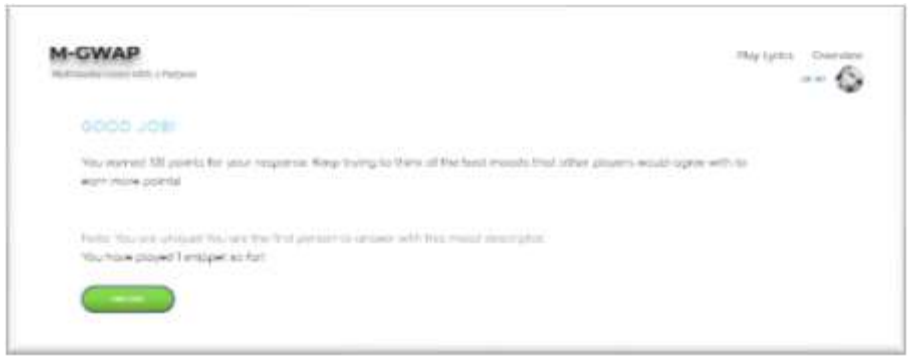

Figure 2: Game results are provided for every answered snippet.

In order to improve the effectiveness of M-GWAP, we implemented additional design features, such as limiting opportunities for cheating, limiting belligerence, encouraging accuracy.

Cheating is a human behavior that is found in various game contexts. In the case of M-GWAP, cheating could occur in two types of behaviors, such as two players sitting next to each other and sharing answers, and a single user playing multiple times the same snippet. In M-GWAP, by providing randomized snippets to the players, two players who sit next to each other will not be receiving the same snippets and their chance to cheat would be very low. The game also never provides duplicate snippets to a player until they responded to all the snippets in the database. This way, a user cannot cheat by playing the same snippet multiple times. Also, snippets do not provide additional points for the most popular annotation, because when players respond to every snippet in the database, they can still play for points without increasing the popularity of mood annotations in the database.

In M-GWAP popularity is here implemented as form of consensus. The game minimizes the negative impact that belligerent players could have on the game by providing junk responses without caring about the impact on the score. Scores computed for emotion labels are based on popularity ranking. Since an emotion term needs to pass a threshold of popularity before it is added as a real annotation for a multimodal excerpt, junk entries remain low in popularity without affecting the data.

The game encourages accuracy by providing more points to players when they give popular mood annotations. This way, players who care about the game and the score are more efficient in their responses to the snippets. By trying to guess the most common annotations and scoring points, the annotations players provide are the most accurate that can be added to the database. Serious players and rewarded and carelessness in playing discouraged. Furthermore, the reward system identified in ranks, leaderboards and badges that users can earn, as discussed, encourages players to aim for high score, ultimately helping M-GWAP to achieve higher accuracy.

While building the design of the game, we encountered the need of creating a game with the least amount of bias in the players' response. Features such as layout, background music, sound effects and graphics could all affect the players' mental state. In designing the game, it was clear that these could also affect the players' perception of the emotions conveyed by the snippets. On the other hand, these features could also help in making M-GWAP more enjoyable to players. Paramount aim of the design was therefore finding a balance that could maintain a high level of interest and not biasing the responses. In order to keep such a balance, we adopted a very simple style of design, including limited use of elements and color, simple icons for badges and well-known gravatar profile pictures as well as the opportunity to upload a personal image for the avatar. While these are features that appeal most players, the simple aesthetics allows the players' focusing their attention on the true purpose of the game, that is guessing the mood of the snippet at hand, not the mood of the game.

Implementing M-GWAP in WordPress encompasses the integration of diverse elements and support different types of devices including mobile. Integration has included a social media infrastructure for user membership support, registration and security features to guarantee the privacy of personal information such as email address, age and spoken languages as well as profile privacy options for users who prefer not to appear in the leaderboard or members page. Ad-hoc implementations included PHP scripts communicating with MySQL through in-page queries. These queries are run in PHP to the MySQL database and use browser sessions, in order to guarantee the security of the data and avoid exploiting vulnerabilities that could exist in WordPress.

## V. M-GWAP EVALUATION

In order to evaluate M-GWAP, we first realized a dataset of text snippets to test the game dynamics described in the present article. For this reason, we realized the dataset from opera lyrics. Most operas are available in the public domain and have expressive lyrics. Opera lyrics are therefore easy to retrieve and useful from the perspective of testing M-GWAP with text snippets that can convey an emotional content having reference to music.

For evaluating the game dynamics from a user perspective (Schell, 2008), we consider M-GWAP both as a videogame and a game with a purpose. Most known approaches for evaluating the potential success of a videogame and the effectiveness of a GWAP in solving research tasks, include the playtest method (Fullerton, 2008; Schell, 2008; Kunkel, Loepp and Ziegler, 2018) and so-called expected contribution (Von Ahn & Dabbish, 2008). In playtest methods, users are questioned about the game directly through both general and specific questions. Through such a survey approach, users can provide direct feedback while also being specific about flaws, usability and enjoyment of a game. Expected contribution is a method measuring the quality of a GWAP by multiplying the average number of problem instances solved per human-hour (throughput) by the average overall amount of time the game will be played by an individual player (Von Ahn & Dabbish, 2008). This method has similarity to testing variables successes, total time and accuracy (Garmen et al., 2019).

**Alpha and Beta testing**

In order to improve our current game before the public release, we carried out a series of alphas and beta tests within the members of Computation of Language Laboratory at University of California, Irvine and some externals. We tested M-GWAP with 33 users and gathered 457 different mood labels for a test dataset of 471 snippets. We count 715 annotations and 185 label and user association, meaning that each label has received 1.5 responses average from 5.6 users average. We report 652 surveys answered reporting mood of the annotator before playing the game. 43 badges were assigned in total, meaning that 1.3 badges were assigned per user on an average. 53 guest users played the game without registering. In these testing, we used only text labels in order to focusing the testing on the game dynamics, as these applies to all media types identically, and to reduce data sparseness. We tested the different media modes separately with a restricted number of selected users. As shown in Table 4, general questions were presented in both alpha and beta tests, together with short-response questions for players' complaints, ideas, clarifications and bug reports.

Table 4. Alpha and Beta testing results.

| Survey's questions | Alpha test (average) | | | Beta test (average) | | |
|---|---|---|---|---|---|---|
| *How enjoyable was the game?* | 4.5 out of 10 | | | 6.5 out of 10 | | |
| *How easy was to navigate through the game and website?* | 8 out of 10 | | | 9 out of 10 | | |
| *How difficult was it to think of a mood for each snippet?* | 3 out of 5 | | | | | |
| *Feedback on the game* | Scoring and feedback is confusing, also after consulting the tutorial (most players) | Game is confusing when tutorial is not consulted beforehand (few players) | Failure in maintaining high engagement; not enough motivation (most players) | No scoring or tutorial problems (all players) | Boredom for snippets (few players) | Lack in quality and quantity of snippets (most players) |

The feedback result from the alpha test allowed us to improve the scoring system. The score ratio and bonus multipliers were adjusted in order to achieve a more balanced scoring approach. In its new version, the scoring system provided 100 base points per player when a new label is provided to the player for a specific snippet, to which 10 extra points are added for every player who already provided the same label. When 100 players provide the same label for a snippet, a multiplier is applied to the score. The multiplier's percentage $m$ varies according to the number of players $p$ who provided the same label that the current player has provided for that snippet, out of the number of players $a$ who have provided any label for the same snippet, multiplied by 1000.

In addition, the alpha testers noticed that the game required providing a better feedback to players regarding the score earned. Alpha testing did not have a clear quality score (i.e. in some games 100 is a good score, which corresponds in this game to the lowest score). We introduced congratulatory words and notes in the results page together with explanations of the score and the number of people who provided the same label already. According to this improvement, when a player earns 100 points, the player also gets educated about the scoring system and informed that a new label has been entered for that snippet. With the aim of providing a game experience as easy and fun as possible, we implemented an in-game tutorial for new players, badges with graphical icons, pictures and avatars for users, and the display of the player's rank compared to others after each game aside of a leaderboard.

These improvements led us to the beta version of the game. As shown in table 4, beta version was more enjoyable and the website easy to navigate. Most problems highlighted by the feedback were fixed. The game requires a larger number of players and snippets to be most enjoyable. Being these problems related to testing stages, the remaining shortcomings are to be expected. In consideration of the improvements carry out following the alpha testing and passing the beta testing successfully, we consider the current version of M-GWAP as ready for release, together with the Musical-Moods dataset, containing more than 10 hours of multimedia, including audio, video, 3D vector based animations from the motion capture data acquired during each performance and text interviews from the participants.

In the current version of M-GWAP, users can play the game in different modalities. Each modality contains either audio, video or text snippets exclusively. For each of the three game modes, separate data tables and user are generated, including scores and leaderboards. On the contrary, the game dynamics are identical for all modalities.

## VI. CONCLUSIONS

GWAPs are games that allows leveraging on crowdsourcing to solve research problems. Because gaming is based on engagement, it is essential for a GWAP to be fun to play. In the present article, we focus on developing a new GWAP for exploiting the wisdom of crowds phenomenon in order to annotate multimedia data in terms of mental states and multiple emotions which can be induced through music. M-GWAP is a multimodal game with a purpose of this kind, drawing from most effective game-structure templates for GWAPS. Because wisdom of the crowds is a popular topic in research and a growing base of users would take advantage from a GWAP easy to implement also with no programming skills, M-GWAP is developed in WordPress. It implements various motivational strategies for the player to remain engaged, such as a score system, text motivators while playing, a ranking system to foster competition and mechanics for identify building. Notably, the present score system is based on thresholds of consensus among a population of annotators that we previously identified in research. We run alpha and beta testing with different user groups, collected feedback and refined the game accordingly.

In the next version of M-GWAP a combined mode will be made available for users to play, where snippets of different media types are presented to the user for annotation simultaneously. In the future, we will modularize M-GWAP into simple and ready-to-use WordPress modules for end users who want to customize M-GWAP to specific needs without having to write a single line of code.

## FUNDING ACKNOWLEDGEMENT

The research is supported by the EU through the MUSICAL-MOODS project funded by the Marie Sklodowska-Curie Actions Individual Fellowships Global Fellowships (MSCA-IF-GF) of the Horizon 2020 Programme H2020/2014-2020, REA grant agreement n° 659434.